\documentclass[conference]{IEEEtran}
\usepackage{times}

\usepackage[numbers]{natbib}
\usepackage{multicol}
\usepackage[bookmarks=true]{hyperref}
\usepackage{amsmath}
\usepackage{amsfonts}
\usepackage{tabularx}
\usepackage[dvipsnames]{xcolor}
\usepackage[pdftex]{graphicx}
\usepackage{multirow}
\usepackage{booktabs}
\usepackage{subcaption}
\usepackage{amssymb}
\usepackage{pifont}
\newcommand{\cmark}{\textcolor{OliveGreen}{\ding{51}}}
\newcommand{\xmark}{\textcolor{Maroon}{\ding{55}}}%
\usepackage{lipsum}

\usepackage[noend]{algpseudocode}
\usepackage[ruled,vlined, linesnumbered, noend]{algorithm2e}

\newcommand{\sayit}[1]{}

\begin{document}

\title{Specification-Guided Data Aggregation for Semantically Aware Imitation Learning}




%
\author{\authorblockN{Ameesh Shah\authorrefmark{1},
Jon DeCastro\authorrefmark{2},
John Gideon\authorrefmark{2}, 
Beyazit Yalcinkaya\authorrefmark{1},
Guy Rosman\authorrefmark{2},
Sanjit A. Seshia\authorrefmark{1}}
\authorblockA{\authorrefmark{1}UC Berkeley \authorrefmark{2}Toyota Research Institute}
\authorblockA{Correspondence to: \texttt{ameesh@berkeley.edu} }}

\maketitle

\begin{abstract}
Advancements in simulation and formal methods-guided \textit{environment sampling} have enabled the rigorous evaluation of machine learning models in a number of safety-critical scenarios, such as autonomous driving. 
Application of these environment sampling techniques towards improving the learned models themselves has yet to be fully exploited. 
In this work, we introduce a novel method for improving imitation-learned models in a \textit{semantically aware} fashion by leveraging specification-guided sampling techniques as a means of aggregating expert data in new environments. Specifically, we create a set of formal specifications as a means of partitioning the space of possible environments into semantically similar regions, and identify elements of this partition where our learned imitation behaves most differently from the expert. We then aggregate expert data on environments in these identified regions, leading to more accurate imitation of the expert's behavior semantics. 
We instantiate our approach in a series of experiments in the CARLA driving simulator, and demonstrate that our approach leads to models that are more accurate than those learned with other environment sampling methods.

\end{abstract}

\IEEEpeerreviewmaketitle
\section{Introduction}
An extensive body of work has demonstrated the effectiveness of \textit{imitation learning} (IL) in a variety of end-to-end control and motion prediction tasks, particularly in the domain of autonomous driving~\cite{bansal2018chauffeurnet, kuefler2017imitating, codevilla2018end, chen2019deep, kebria2019deep}. IL methods, which learn a policy to mimic collected samples of expert behavior, are typically highly general and can be easily implemented in a variety of domains. Despite these successes, IL methods often suffer in performance due to the problem of \textit{covariate shift}: inaccuracies in predicted actions early in a control episode may lead to compounding error when visiting future states outside the original training distribution~\cite{spencer2021ilfeedback}. In this work, we propose a novel method to combat covariate shift that leverages configurable environment domains, such as simulations, to aggregate expert data in \textit{varied environments} from which an accurate imitation can be learned. 

Existing approaches to reduce error caused by covariate shift, namely DAgger-style methods~\cite{ross2011dagger, prakash2020daggerplus, cui2019uail, kelly2019hgdagger, zhang2017safedagger}, propose switching control between an IL agent and the expert policy in order to collect expert data on states that were outside the original training distribution. This family of algorithms generally focuses on \textit{when} to switch control and collect data from the expert, by exploiting additional objectives or the uncertainty of the learning model. We devise an approach that chooses \textit{where} the agent will act in its world, as the world in these methods is usually treated as fixed, not a controllable part of the IL loop.

As a motivating example, consider the task of imitating a student driver to better understand how and when the student may make errors during driving. Since the student is a novice, we want to avoid putting the student in a real vehicle and instead opt to have the student operate in simulation. Further, since our goal is to imitate the student as accurately as possible, we are not explicitly seeking to train the safest or most performant learned model. If the student makes a mistake in a particular circumstance, we want our learned imitation to make the same mistake as well.

With a controllable world, the key question becomes: How do we find environments in this world that will help us broaden our understanding of the expert's behavior to yield a better imitation? If the student driver typically drives very hesitantly and avoids getting close to other vehicles, how would the student behave if they were to be suddenly cut off by another car? In domains like driving, unlikely events (such as cutoffs) have high \textit{semantic value} for both the student and teacher (collisions can occur as a result of cutoffs.) When learning an imitation, it is crucial we prioritize these circumstances and find environments that induce such outcomes so that we may collect valuable expert data.

In order to find meaningful environments, we introduce an approach that semantically describes and differentiates amongst the many possible environments in our world, leveraging domain-relevant information in the form of logical properties. In the case of our driving example, one such property could capture whether the student driver encountered a collision, and another could measure whether a speed limit was exceeded. From these properties, we create a set of \textit{formal specifications} by enumerating all of the possible truth value assignments to these properties. Each specification corresponds to \textit{all} possible outcomes that assignment could capture. Following our example, these specifications would separate environments where the student encountered a collision but didn't speed, ones where the student sped but didn't collide, ones where the student both collided and sped, and ones where the student neither collided nor sped.

This set of specifications effectively \textit{partitions} the space of possible (environment, agent trajectory) pairs in our world: a single pair can only satisfy (i.e., belong to) one specification in our set and no other. This partition offers a semantic characterization of the space of environments that we can then use to guide the data aggregation process.

\begin{figure*}
     \centering
\includegraphics[width=.98\linewidth]{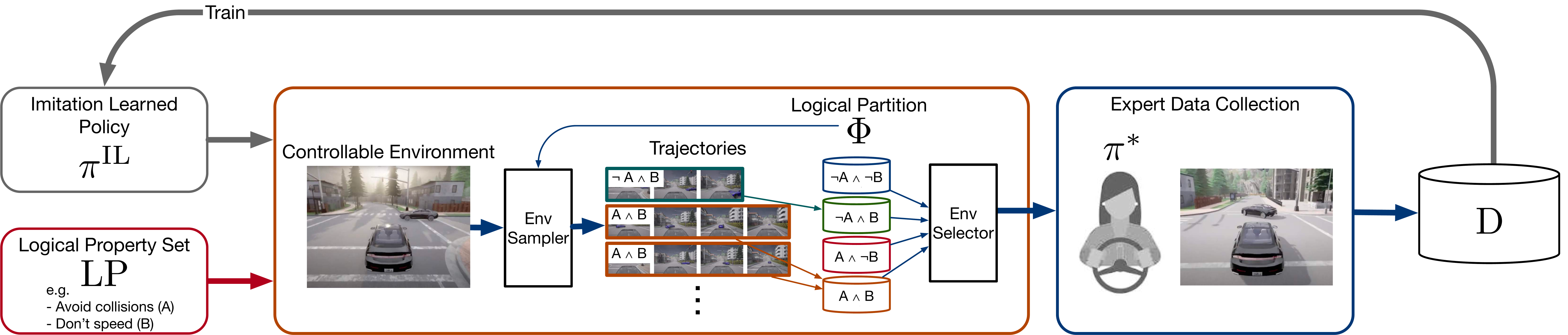}
     \caption{Our approach for environment sampling for data aggregation. Using a set of logical properties, we sample varied environments and select ones that induce semantically different behavior between our expert and an IL policy.}
    \label{fig:overview}

\end{figure*}

We provide a visualization of our data aggregation algorithm in Fig.~\ref{fig:overview}. We deploy our current IL policy in various environments, using the aforementioned partition to equitably sample from each specification and gather a semantically diverse set of environments. From this set, we select a subset of environments where we most expect the expert and IL policies to \textit{disagree} on their semantic behaviors, and aggregate expert data on this subset.

We concretely instantiate our approach in an algorithm we call \textit{Specification-Guided Data Aggregation} (SGDA). We demonstrate the efficacy of our algorithm in the CARLA autonomous driving simulator by imitating two different expert models as they perform different driving maneuvers at a four-way intersection. Our results show that aggregating data using environments selected by our algorithm leads to better IL accuracy than environments selected by baseline methods, especially in unlikely but semantically meaningful circumstances. 

In summary, our contributions in this paper are threefold: First, we identify a novel formalism -- a semantic partitioning of possible (environment, trajectory) pairs based on the formal specifications the trajectories satisfy -- for systematizing environment sampling. Second, we propose an algorithm that uses the aforementioned partitioning to improve imitation learning models in a data-aggregation loop. Third, we show promising experimental results of our algorithm for human driver modeling, and improved performance over environment sampling baselines in the context of autonomous driving.

\section{Related Work}





\textbf{Robustifying Imitation Learning.}
Several works address the problem of robustness in imitation learning, either through modification of the training criteria or extending to environments that cover properties of interest, such as rare events.  \cite{codevilla2019} acknowledge and show that dataset bias can hinder achieving robustness to corner cases, and propose algorithmic improvements and auxiliary training tasks.  When a task is known, \cite{wang2017, Zolna2019TaskRelevantAI, bhattacharyya2020humangail} all use modified generative adversarial imitation learning (GAIL) structures to learn semantic policy embeddings and enforce task-relevancy.  \cite{lu2022imitation} augment pure IL with a reinforcement learning (RL) agent to deal with distribution shifts.  Other works have explored using confidence measures to characterize trajectories and select among them for training.  \cite{zhang2021confidence, Huang2019-ce} use notions of confidence to align imitation with downstream tasks, while \cite{Richter2017-lh, Amini2018-vl} use notions of novelty detections trained on diverse datasets.

We also address the robustness problem, but view it from the perspective of aggregating expert data to ensure a desired level of IL performance can be achieved.  Decoupling from the IL approach eliminates constraints on the learning pipeline and frees users to adopt any IL approach.



\textbf{Falsification and Importance Sampling.}
Several works use formal methods and search-based methods to uncover rare events and property violations in safety-critical systems.  Some handle the \textit{falsification} problem, including \cite{abbas2014conformance}, who propose an approach that gauges conformance of a given model with respect to a reference model for which certain guarantees hold, while \cite{akazaki2018falsification, lee2020} use reinforcement learning to reduce the number of simulations required to find counterexamples. \cite{chou2018using} cast the falsification problem as one of control synthesis, in which invariant sets are used as the basis to concentrate sampling.  Some mature falsification approaches are implemented in software tools such as s-Taliro~\cite{annpureddy2011s} and Verifai~\cite{dreossi2019verifai}.

In contrast to falsification, \textit{importance-sampling} (IS) methods have shown promise in finding counterexamples to desired properties.  Works such as \cite{sankaranarayanan2012falsification} adapt the search for violations of a property using the cross-entropy method, assuming a fixed partitioning of the input space.  
\cite{AriefHKBHDLZ21} extend standard IS by training a deep-learned sampler with efficiency guarantees, while \cite{OKellySNTD18} adopt an adaptive IS procedure with learned distributions over the search space to further accelerate discovery of rare events.  Because many such approaches quickly hone in on rare events, it remains challenging to directly apply them to the task of supplementing data used for learning, as they may miss many semantically-interesting non-failure outcomes.


\textbf{Scenario Generation for Safe Autonomy.}
While the above works focus on sampling in general, there is a substantial body of literature focused on diversifying the scenarios used for learning safe autonomy systems.  \cite{DingCLEZ21, DingLJZ21} train flow-based generative models and networks exploiting hierarchical structures to sample environments for both training and validation.  Other works incorporate dynamics and simulations of the world to inform sampling~\cite{WengCOR22, gu2022review, lee2020}.
The adaptive sampling works in \cite{OKellySNTD18, WangPTMS0RU21, UKSz19} search over an environment parameter space to rapidly uncover failure modes.
Some works, such as the Scenic language~\cite{scenicOriginal}, have explored more programmatic, user-readable ways of specifying these modes.  While such works perform environment sampling, our approach differs in that we balance our dataset across a semantically-partitioned space, and directly apply sampling methods on training imitation-learned agents.




\textbf{Data Aggregation Techniques.}
Online sampling frameworks for data aggregation such as DAgger~\cite{ross2011dagger} have been employed to address the problem of distribution shift when training imitation learning models.  HG-DAgger~\cite{kelly2019hgdagger} queries a human expert for supplemental corrective actions to improve the robustness of a learner.  Other variants, such as Safe-DAgger~\cite{zhang2017safedagger}, data aggregation for policy learning \cite{prakash2020daggerplus} and uncertainty-aware data aggregation \cite{cui2019uail} all propose query-efficient sampling.  The aims are similar to ours in that their attempts to reduce the data burden from human experts by focusing on safety- and uncertainty-related criteria.  However, in contrast to existing approaches, ours uses task-relevant properties and samples different environments where the learner diverges from the desired property set. We emphasize that our approach can be incorporated into any of the aforementioned approaches.

\section{Methodology}



Given an initial dataset collected from an expert policy, a continuous control task, and a controllable environment, we aim to use logical specifications to learn an accurate imitation of the expert. In this section, we formulate our method of semantically sampling possible conditions of our environment for use in a data aggregation-style loop.

We start by outlining episodic control under a given policy in a controllable environment. Let $\mathcal{E}$ represent \textit{all} possible environment conditions (realizations of the environment) that we can represent in our simulator. Given an environment condition $e \in \mathcal{E}$ and a policy in the class of policies available to the learner, $\pi \in \Pi$, we define a control episode as $ \xi_{\pi}(e) = d(\pi, e)$. Here $\xi_{\pi}(e)$ is a trajectory of state and action pairs of the expert, and $d$ represents a function that rolls out the dynamics for a control episode using a fixed policy $\pi$ to control the agent and $e$ to set the conditions of the environment.  
For simplicity, we will assume that our dynamics are deterministic and that $d$ returns a single trajectory. The function $d$ unrolls the dynamics until the trajectory reaches a predefined termination condition, such as arriving at a goal destination, violating a safety condition, or reaching a maximum number of timesteps. 

In this section, we first describe how to semantically divide the space of possible trajectories by constructing a \textit{semantic trajectory partition} (STP), then describe how we use this partition in a data aggregation-style imitation learning loop.

\subsection{Creating a Semantic Trajectory Partition}
\label{sec:env_partition}
We begin by having a user provide a set of (temporal) \textit{logical properties, LP}, that are domain-specific and identify meaningful properties of an agent's trajectory. For example, in the context of autonomous driving, these properties could represent avoiding collisions or maintaining speed limits. The formalization of such properties are well-explored and can be adapted from works like~\cite{censi2019rulebooks}.
We allow for flexibility in how these properties are expressed: they can be purely Boolean properties (avoid all collisions) expressed in logics like Metric Temporal Logic (MTL)~\cite{MaierhoferMTL2020} and Higher-Order Logic (HOL)~\cite{RizaldiTrafficRules2015}, or a Boolean property accompanied by a continuous objective (stay a thresholded distance away from all other vehicles), expressed in quantitative semantics like Signal Temporal Logic (STL)~\cite{maler2004monitoring, HekmatnejadRSS2019}.
We denote the size of $LP$ as $\vert LP\vert = l$.

With this property set $LP$, we construct a partition of the space of all agent trajectories. Our approach for creating the STP is the following.  We first treat each property $p \in LP$ as a \textit{propositional variable} that is true if and only if a trajectory $\xi_{\pi}(e)$ satisfies $p$ under environment $e$; that is, the pair $(\xi_{\pi}(e), e) \models p$.  Then, we enumerate all $2^l$ possible Boolean logical formulae conjuncting the properties $p$ in $LP$ or their negations. In doing so, we create a set of \textit{logical specifications}, $\Phi$, that covers all possible combinations of truth values for every property in $LP$. We will refer to elements of our partition $\varphi \in \Phi$ as \textit{specifications}. We will often abuse this notation and refer to specifications both as the logical formula associated with a partition element \textit{and} the set of collected trajectories satisfying that logical formula (i.e., exist within the partition element) when aggregating data.

\begin{figure}
    \centering
    \begin{subfigure}{\linewidth}
        \centering
        \includegraphics[width=\linewidth]{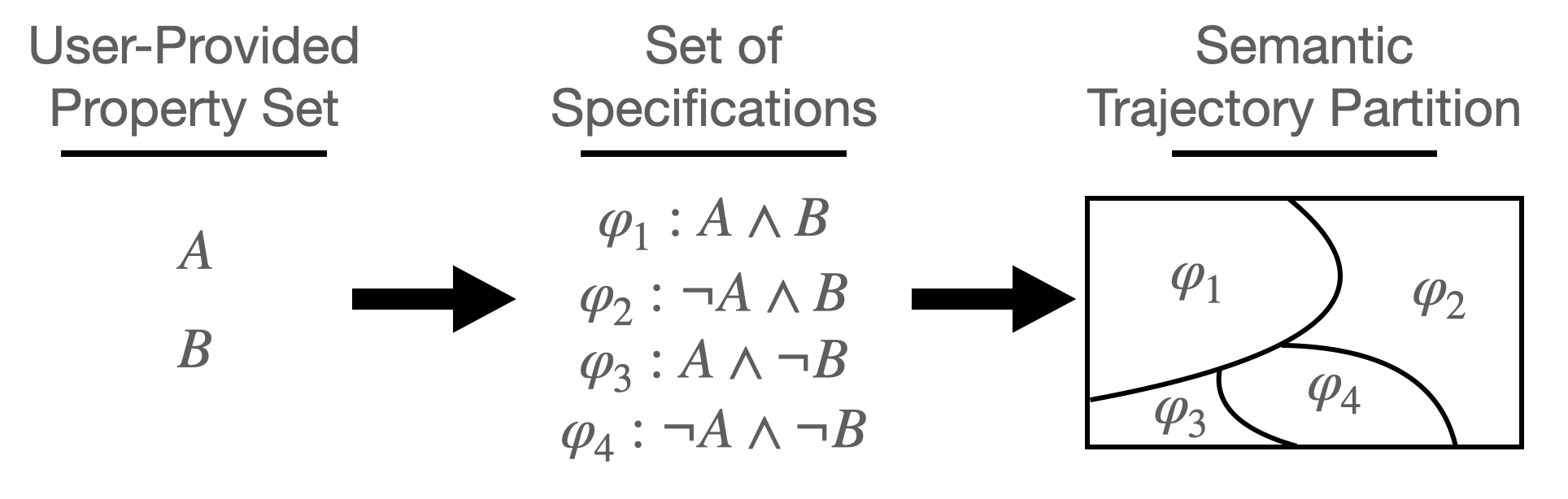}
    \end{subfigure}
    
    \begin{subfigure}{0.245\linewidth}
        \includegraphics[trim={0cm 2.4cm 3.75cm 0cm},clip,width=\linewidth]{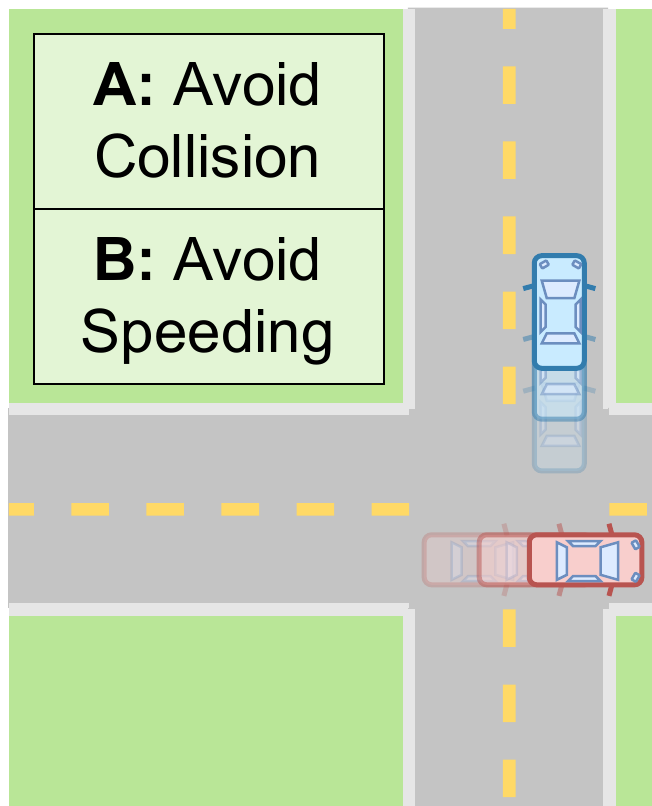}
    \end{subfigure}%
    \hfill%
    \begin{subfigure}{0.245\linewidth}
        \includegraphics[trim={0cm 2.4cm 3.75cm 0cm},clip,width=\linewidth]{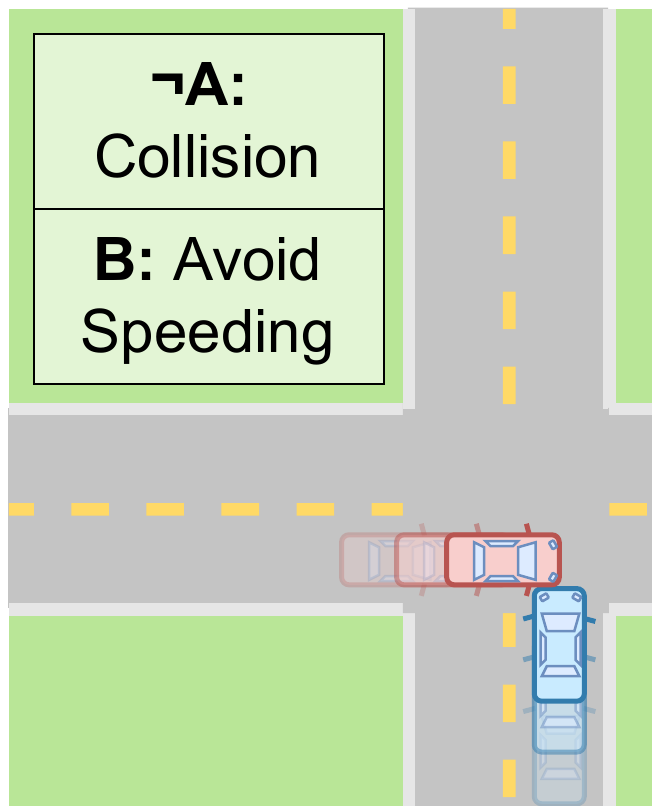}
    \end{subfigure}%
    \hfill%
    \begin{subfigure}{0.245\linewidth}
        \includegraphics[trim={0cm 2.4cm 3.75cm 0cm},clip,width=\linewidth]{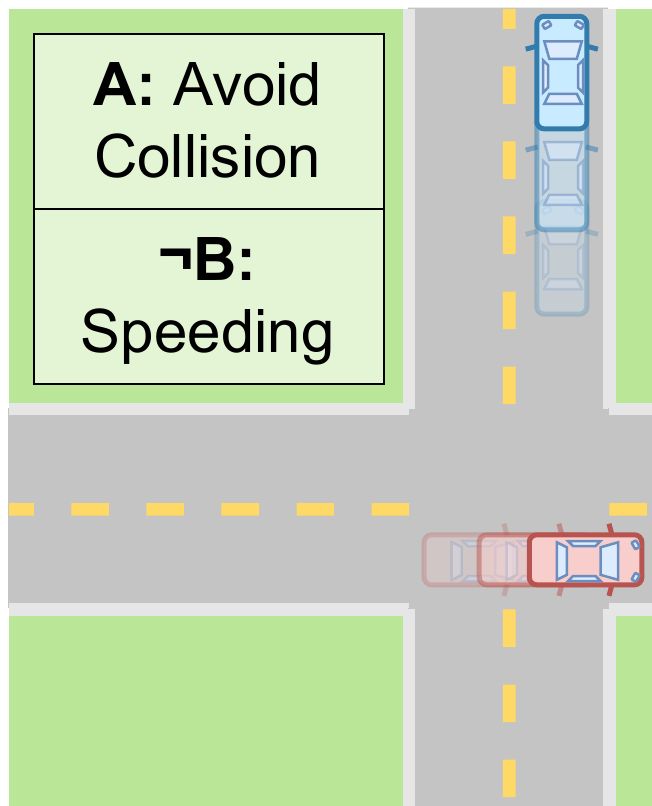}
    \end{subfigure}%
    \hfill%
    \begin{subfigure}{0.245\linewidth}
        \includegraphics[trim={0cm 2.4cm 3.75cm 0cm},clip,width=\linewidth]{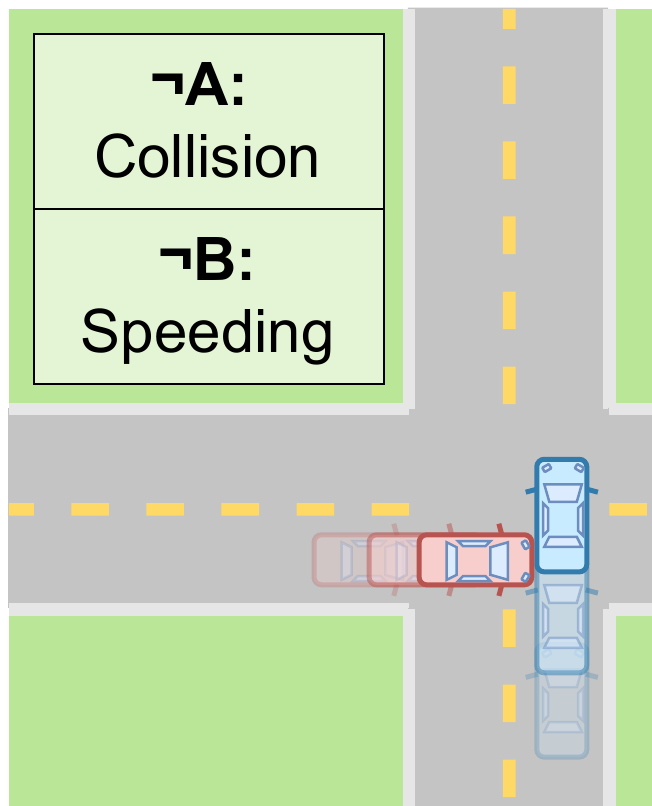}
    \end{subfigure}

    \caption{\textbf{Top:} A visualization of how user-provided properties can be treated as propositional variables to form our STP. We enumerate all combinations of truth values for each property to create a set of specifications that act as elements in a partition of the trajectory space. \textbf{Bottom:} An example of trajectories that satisfy each specification for the blue car agent.}
\label{fig:example_figure}
\end{figure}

As a simple example, consider the scenario depicted in Fig.~\ref{fig:example_figure}. In this example, we have a set $LP$ with just two properties: ($A$) avoid collisions, and ($B$) stay within the speed limit. Our partition $\Phi$ would then consist of four elements: $(A \land B)$ (avoid collisions and don't speed), $(\neg A \land B)$ (collide and don't speed), $(A \land \neg B)$ (don't collide and speed), and $(\neg A \land \neg B)$ (collide and speed.) Note that every trajectory that an agent can produce will satisfy only one of the specifications in our partition. Also, notice that in order to explore the full partition, certain environments the agent is in will elicit more samples in specific elements of $\Phi$ --- e.g., an urban environment that favors slower speeds. 

We note that the size of $\vert \Phi\vert = 2^l$ may limit in practice the number of properties a user can provide. We do not see this as a prohibitive issue, however, primarily due to the fact that users only have to specify a small number of properties in order to sample a diverse set of data, as evidenced by our experiments. Further, our approach can be generalized to a setting where a user can assign importance weights to each property, which can allow for the de-emphasis of specific partition elements. We further expand on this generalization in the appendix.  

During training, we can use our STP to sample equitably from the possible space of semantic results and ensure there is ample coverage of these semantics in the data. We will next discuss how to use the STP and its elements 
to improve imitation learners via data aggregation.

\subsection{Specification-Guided Data Aggregation}\label{sec:spec_guided_data_agg}

Ultimately, our goal is to use $\Phi$ to improve the accuracy of our imitation learning agent $\pi^{\text{IL}}$ by aggregating a semantically diverse dataset. To do this, we take an approach consisting of four steps: (1) Training, (2) Environment Sampling, (3) Environment Selecting, and (4) Data Aggregation, as outlined in Algorithm~\ref{sgda_alg}. We give an explanation of each step in turn:

 \begin{algorithm}
\SetAlgoLined
\KwIn{Initial dataset $D$, Number of aggregation rounds $n$, STP $\Phi$, expert policy $\pi^*$, \# of samples $k$, \# of selections $m$}
 Learn $\pi^{\text{IL}}$ w.r.t $D$\;
 Initialize $E_\text{prev}, D_\text{prev}$ to $\emptyset$\;
 \ForEach{iteration $i \in 0 \dots n$}{
  $E_\text{new} := \text{EC-Sampling}(\pi^{\text{IL}}, k)$\; 
  $E_\text{new} := \text{EC-Select}(E_\text{new}, E_\text{prev}, D_\text{prev}, m)$\;
  $D_\text{new} := \{d(\pi^*, e); \forall (e, \xi_{\pi^{\text{IL}}}(e)) \in E_\text{new}\}$\;
  $D = D \cup D_\text{new}$; \tcp{aggregate} 
  Learn $\pi^{\text{IL}}$ w.r.t $D$\;
  Set $E_\text{prev}, D_\text{prev}$ to $E_\text{new}, D_\text{new}$\;
  }
 \Return{$\pi^{\text{IL}}$}
 \caption{Specification-Guided Data Aggregation (SGDA)}
     \label{sgda_alg}
 \end{algorithm}

\textbf{Training.} To start, we assume an initial dataset 
$D$ collected from an expert. We make no assumptions on the distribution of the environment conditions from which $D$ was collected. We then learn an imitative policy, $\pi^\text{IL}$ from $D$ using any IL algorithm, such as Behavioral Cloning (BC) or GAIL~\cite{ho2016gail}. 

\textbf{Environment Sampling (EC-Sampling).} \label{sec:env_sampling}
In this step, we attempt to collect a semantically broad pool of trajectories by sampling a set of environment conditions $e \in E$ and executing $ \xi_{\pi^\text{IL}} = d(\pi^\text{IL}, e) $. 

When sampling and collecting different environment conditions, we balance two factors: First, we want to collect trajectories for every element of our partition $\Phi$ as equitably as possible. Second, we do not want to attempt to sample for elements in the partition that are very hard to satisfy (i.e., we attempt to sample environment conditions that produce a trajectory in a given partition element $\varphi$, but finding the correct environment conditions is challenging). To find the correct environment conditions for a given $\varphi$, we leverage optimization-based sampling methods, such as Bayesian Optimization or Cross-Entropy, which learn a distribution from already sampled (environment condition, trajectory) pairs that maximizes the chance of a successive sample satisfying our target specification $\varphi$. 

In order to balance the aforementioned two factors, we use the Upper Confidence Bound algorithm~\cite{auer02ucb} to select the next target specification we sample for, adopting notation from~\cite{khandelwal2016mctsucb}. The UCB algorithm balances \textit{exploitation} (sample for partition elements that are under-represented) and \textit{exploration} (don't sample for partition elements that we have made a lot of sample attempts for.) 

We outline our approach in Algorithm~\ref{ecs_alg}. We keep a total count of how many times we sample for a selection with $N$, and a count of how many $(\xi_{\pi}(e), e)$ pairs satisfy each specification with count (line 2). In each iteration, we choose a target specification $\varphi_c$ that we will attempt to sample a satisfying environment condition for. In lines 5-6, we increment the count for $\varphi_c$ and sample for it using our optimization-based sampler, which we denote as $f$. $f$ takes in as arguments the target specification and the collected samples and produces a new sample environment condition that the optimizer believes is likely to satisfy $\varphi_c$. In lines 7-9, we roll out our dynamics and collect the trajectory corresponding to the sampled environment, and add that trajectory to the element partition that it actually falls under by checking satisfaction for every specification in our set. In lines 10-11, we compute the \textit{value} of choosing a specification for the next iteration by prioritizing specifications that have been undersampled relative to other specifications. We use these values along with the number of times we attempted to sample for a given spec ($N$ in our algorithm) in the UCB computation in line 13 to choose the specification to sample for in the next round.

\begin{algorithm}
\SetAlgoLined
\KwIn{Environment Partition $\Phi$, Optimization-guided sampler $f$, learned policy $\pi^{\text{IL}}$, Number of iterations $k$}
 Select an initial $\varphi_c$ from $\Phi$\;
 Initialize $N_{\varphi_j}$ to 0 $\forall \varphi_j \in \Phi$\; 
 Initialize $S$ to $\emptyset$\;
 \ForEach{iteration $t \in 0 \dots k$}{
  $e_t := f(\varphi_c, S)$; \tcp{get new sample} 
  Increment $N_{\varphi_c}$\;
  $ \xi_{\pi^{\text{IL}}}(e_t) = d(\pi^{\text{IL}}, e_t) $; \tcp{roll out dynamics}
  Add $\xi_{\pi^{\text{IL}}}(e_t)$ to the $\varphi$ it satisfies\;
  \ForEach{$\varphi_j \in \Phi$}{
    Set $Q_j$ to $max_{\varphi \in \Phi}(|\varphi|) - |\varphi_j|$\;
  }
  Add $(e_t, \xi_{\pi^{\text{IL}}}(e_t))$ to $S$\;
  Set $\varphi_c := \text{UCB}(Q, N, t)$ \tcp{pick next spec}} 
 \Return{S}
 \caption{Environment Condition Sampling (EC-Sampling)}
     \label{ecs_alg}
 \end{algorithm}

\textbf{Environment Selection (EC-Select).} With our collected set of environment conditions and trajectories from the previous step, we want to select a subset of conditions for the next round of data aggregation that will improve our model in a semantically meaningful way. For this, we leverage the following insight: Environment conditions $e$ that lead to different satisfied specifications between our learned policy's trajectory and our expert policy's trajectory should be prioritized higher when selecting. Formally, we want to bias toward environments where, when rolled out with our expert and imitative policies, lead to $(\xi_{\pi^*}(e), e) \models \varphi_i$ and $(\xi_{\pi^{\text{IL}}}(e), e) \models \varphi_j$, respectively, where $\varphi_i \neq \varphi_j$.

In our initial round of data aggregation, we do not have a set of environment conditions with trajectories from both our learned model and the expert model. In this case, we simply sample uniformly at random from our set to select our dataset for aggregation. In future iterations, however, we have the previous iteration's dataset, containing both the collected expert trajectories $(\xi_{\pi^*}(e), e)$ and our learned model's trajectories $(\xi_{\pi^{\text{IL}}}(e), e)$ for many $e$. We can then compare whether the same environment condition led to the same specification being satisfied by both the expert and learned model trajectories. 

\begin{table}[]
    \centering
    \caption{Example computation for environment selection weights.}
    \label{tab:selection_example}
    \begin{tabular}{c|cccc}
        \toprule
        & \multicolumn{4}{c}{IL outcomes} \\
        Expert outcomes & $A \land B$ & $\neg A \land B$ & $A \land \neg B$ & $\neg A \land \neg B$ \\
        \midrule
        $A \land B$: 10 & 8 & 2 & 0 & 0\\
        $\neg A \land \neg B$: 10 & 0 & 0 & 5 & 5 \\
        \midrule
        Selection weights & 0.2 & 1 & 1 & 0.5 \\
        \bottomrule
    \end{tabular}
\end{table}

As an example, consider the two-property partition introduced in Section \ref{sec:env_partition}. Suppose in the previous round of data aggregation, we selected a set of 20 environment conditions. In this hypothetical scenario, our expert avoided collisions and abided by the speed limit $(A \land B)$ in 10 conditions, and collided and violated the speed limit $(\neg A \land \neg B)$ in the other 10 conditions. In the same 10 conditions that led to $(A \land B)$ from the expert, our learned model avoided collisions and abided by the speed limit in eight of them, and in the other two, collided while abiding by the speed limit $(\neg A \land B)$. In the 10 conditions that led to $(\neg A \land \neg B)$ from the expert, our learned model also collided and violated the speed limit in only five of them, and in the other five, didn't collide but violated the speed limit $(A \land \neg B)$. This scenario is outlined in Table~\ref{tab:selection_example}. We would set up our \textit{selection weights} as:
\[
1 -(M_{\varphi} / N_{\varphi}) \quad \text{if} \quad N_{\varphi} > 0, \quad \text{else} \quad 1
\]
 
Here, $N_{\varphi}$ refers to the total number of instances in which a specification was satisfied by either the expert or imitation (without double counting for both), and $M_{\varphi}$ refers to the number of expert outcomes that were matched by the imitation learner. 
These weights are normalized into a categorical distribution, and we sample from the pool of sampled environment conditions (from our previous step) weighted by this distribution to yield our selected batch for the subsequent iteration of data aggregation.
We denote this selection step by $\textrm{EC-Select}(E_\text{new}, E_\text{prev}, D_\text{prev}, m)$ in Algorithm~\ref{sgda_alg}, using the previous expert and imitated data to compute selection weights and select $m$ samples from the new set $E_{\text{new}}$.

\textbf{Data Aggregation.} With a final selected set of environment conditions, we can collect trajectories from the expert policy under these conditions and aggregate the collected data in our overall dataset. This dataset can then be used to train our imitation learned model in the subsequent iteration; that is, we repeat the process with our training step on this aggregated dataset. 

Algorithm~\ref{sgda_alg} summarizes the process above; it first trains an initial imitation (line 1), then selects new environment conditions using the aforementioned environment sampling and selection methods (lines 4-5), and queries the expert with the conditions (line 6) to produce a dataset that can be aggregated (line 7). The process can be repeated as desired.

\textbf{Overhead of SGDA.} When actuating dynamics for each $e$ and evaluating satisfaction for each $(\xi_{\pi}(e), e) \models \varphi$, SGDA incurs computational overhead relative to approaches that sample environments without any guidance. We find that this overhead scales linearly with the size of $\Phi$, and note that this trade-off in practice is often much smaller in comparison to the high cost of expert-in-the-loop data collection, which SGDA aims to reduce. We provide a more detailed analysis of this overhead in the appendix.
\section{Experimental Setup}

We evaluate our method on the task of imitating an expert in an autonomous driving task. Below, we provide details on our simulator setup, baselines, and metrics for evaluation.

\subsection{Simulator and Experimental Design}
We use the CARLA simulator~\cite{coilICRA2018}, a high-fidelity autonomous driving simulator that allows for the adjustment of a number of environmental factors, such as the positions and behaviors of other agents in the environment, the layout of roads, and the presence of obstacles. For our experimental domain, we consider imitating an expert's behavior in a four-way intersection, where the expert, serving as the ego vehicle, proceeds through the intersection to reach a destination.

\begin{figure}
     \centering
\includegraphics[width=.8\linewidth]{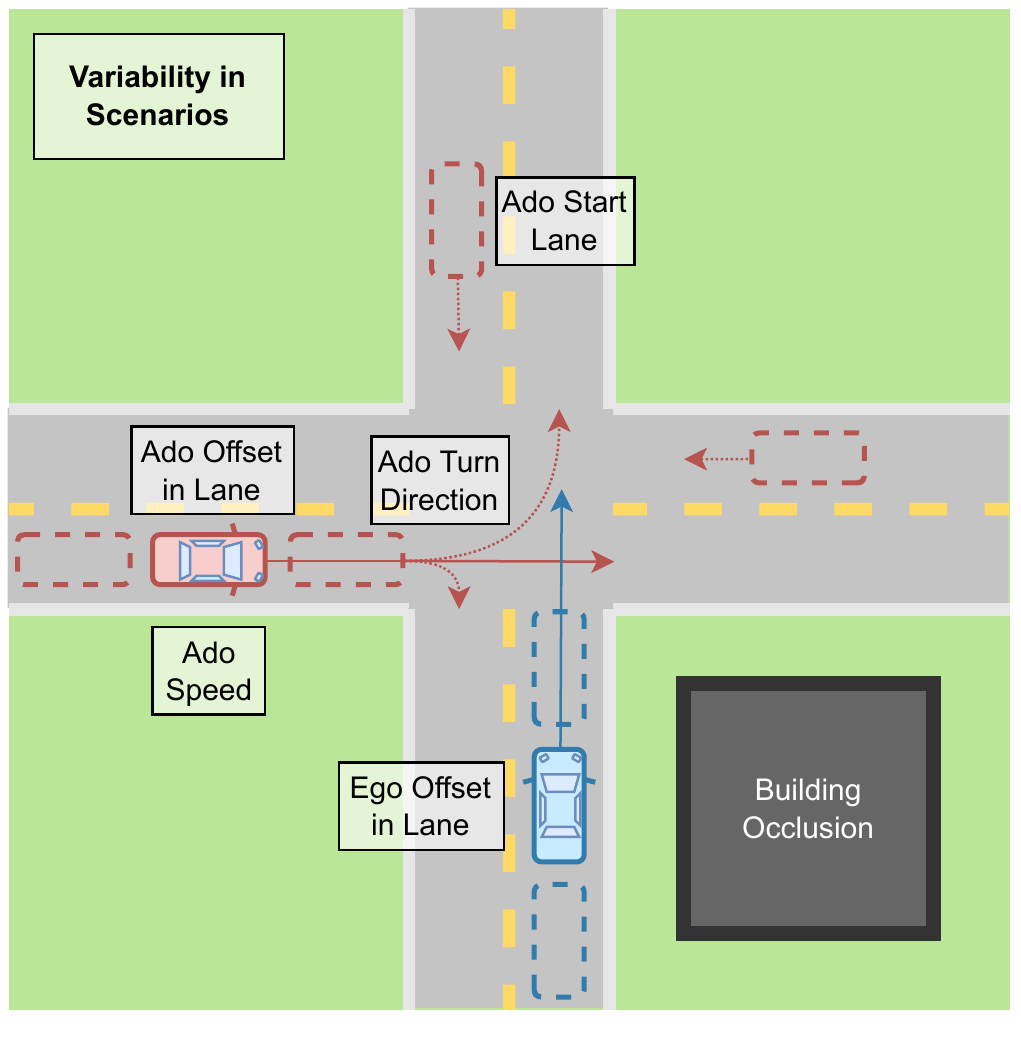}
     \caption{Our experimental setup in the CARLA simulator. We outline the environment parameters that can vary our scenario.}
     \label{fig:experimental_domain}
\end{figure}

In this setting, another vehicle (the \emph{ado}) may be present and rushing through the intersection, ignoring traffic laws.
The ego's trajectory will be affected by the conditions of the intersection environment, which we define as the following parameterized variables: First, the ado vehicle may be spawned at any other side of the intersection (either opposite the ego or to either side of the ego), within a range of varying distances from the intersection. The ado may also perform one of three maneuvers at the intersection: a right turn, a left turn, or proceed straight.
The ego's initial distance from the intersection, as well as the ado vehicle's maximum and minimum speeds, are also variable environment condition parameters. The ado's behavior is controlled by a simple programmatic controller that accelerates the ado to its maximum speed as it traverses the intersection. We vary all of our parameters in order to sample new environments in our data aggregation procedure. Our experimental setup is presented in Fig.~\ref{fig:experimental_domain}.

\vspace{-0.3mm}
We define a set of properties in Signal Temporal Logic (STL), each with its own quantitative semantics to measure how close to satisfaction each property is. Our property set, $LP$, for our experiments are as follows:
\begin{enumerate}
    \item \textbf{Avoid collisions.} This property is falsified if we detect the ego vehicle has collided with any object in the environment. We define the quantitative semantics of this property as the minimum distance between the ego vehicle and the ado vehicle during a trajectory. We treat all collisions as the same.
    \item \textbf{Do not halt.} This property is falsified if the ego vehicle comes to a complete stop during the trajectory. The quantitative semantics of this property are measured by the minimum speed the ego reaches during a trajectory.
    \item \textbf{Do not abruptly brake.} This property is falsified if the ego vehicle brakes with an intensity beyond a certain pre-defined threshold (set to 40 percent of the ego's maximum braking capacity). The quantitative semantics of this property are measured by the maximum braking value the ego applies during a trajectory. 
\end{enumerate}

We compare SGDA with the following environment sampling methods as baselines:

\begin{enumerate}
    \item \textbf{Uniform Prior Sampling}: In order to aggregate more data from the expert, we sample from the environment's prior (for which we set all parameters to be a uniform distribution) and use these samples to collect additional trajectories from the expert for training in each round.
    \item \textbf{Single Specification-Falsification}: Rather than using our STP to sample environments, we instead define a single specification that indicates an outcome of interest and sample for environment conditions that falsify this specification. We set this specification to be the conjunction of all the original (non-negated) properties listed above, which describes `typical' driving behavior for the ego.
    \item \textbf{Individual Property Importance Sampling}: Without the partition created in our SGDA method, an alternative that makes use of all properties in our set would be to sample for the falsification of each property individually (independently of one another). We split up the number of environments to aggregate by the number of properties and importance sample for each, taking an approach for each property similar to that used in~\cite{OKellySNTD18} with a fixed budget of samples.
\end{enumerate}

We measure the efficacy of our approach through a set of quantitative metrics that measure both how \textit{semantically} similar our imitations are to the expert, as well as how \textit{physically} similar expert and imitation trajectories are: 
\begin{enumerate}
    \item \textbf{Outcome Matching:} Similar to the environment selection method described in Section~\ref{sec:spec_guided_data_agg}, we measure how often the expert’s trajectory satisfies the same specification, or \emph{outcome}, as our learned model's trajectory under the same $e$.
    \item \textbf{Trajectory-Length Distance:} 
    To measure trajectory-level differences, we use Dynamic Time Warping (DTW) distance to compare trajectories between our expert and our learned models under the same $e$.
    \item \textbf{Test Set Loss:} 
    We use an L1 Loss over a test set of states and actions collected from our expert to measure how our learned models' actions compare to that of the expert under the same states.
\end{enumerate}

We evaluate our method by imitating two different experts: the CARLA built-in autopilot vehicle controller (AP), and a neural-network controller trained on data collected from a human driver who operated in our environment (NN). The neural-network expert controller was trained on two hours of data collected by having a human operate a steering wheel and pedal to drive the ego car in simulation through the intersection repeatedly. The human was instructed to proceed through the intersection as quickly as possible while avoiding collisions, if possible. For the autopilot, we conducted two additional experiments for two additional maneuvers: taking a right turn or a left turn at the intersection. In the case of the autopilot right- and left-turn experiments, we only learn an imitation for the throttle component of the maneuver and let the expert policy control the steering to avoid obscuring the results of imitation with simultaneously acquiring a challenging lateral control skill in a data-constrained setting. 

We use the Behavioral Cloning (BC) algorithm to learn an imitative policy of the expert, where the state space consists of the ego vehicle's information (location, velocity, acceleration), and a noisy estimate of the ado's location and speed if the ado is within the line of sight from the ego vehicle.

 We collect an initial expert dataset of 200 trajectories (which corresponds to roughly 90,000 state-action pairs on average), with environment conditions sampled from the simulator's uniform prior over all environment parameters. Subsequent rounds of data aggregation are 100 episodes each. In the cases of SGDA, the Single-Specification baseline, and the Individual Property Sampling baseline, 200 samples are generated in the sampling step, and 100 samples are selected to use for aggregation. Of these 200 samples, the first 100 are collected uniformly at random to seed our optimization-guided falsifier(s) for each specification. In the case of SGDA, the subsequent 100 samples are collected using Algorithm~\ref{ecs_alg}. In our baselines, those 100 samples are either used to falsify the specification (in our Single-Specification baseline) or equally divided amongst each of our properties and importance sampled (in the case of the Individual-Properties baseline). Our optimization-guided falsifiers use Bayesian Optimization to learn a distribution over the environment parameters, where the target value provided is the quantitative evaluation of the STL specification the optimizer aims to falsify. On top of this target value, we add a term to incentivize exploration and prevent the distribution from centering too much probability mass on a small region of the parameter space (details in the appendix). We report results after two rounds of data aggregation for each method.

\section{Results}
\begin{figure}
     \centering
\includegraphics[width=0.8\linewidth]{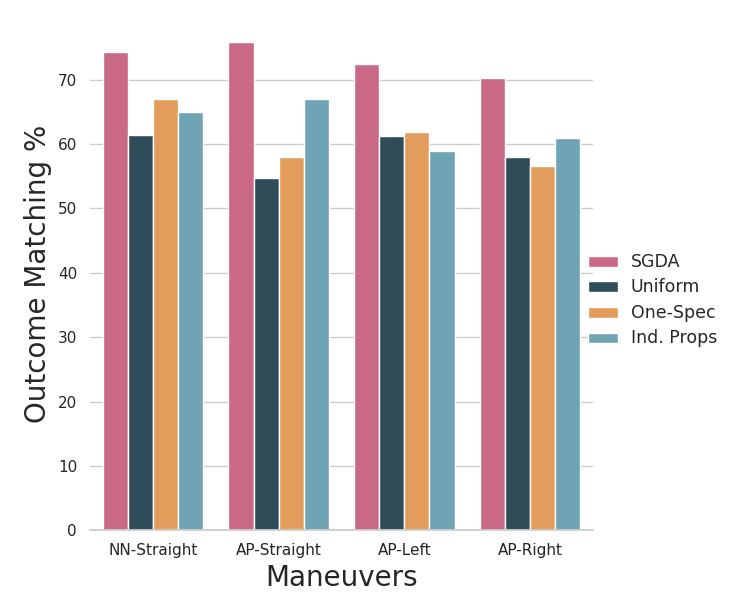}
     \caption{Results for Outcome Matching rates for imitations of the maneuvers performed by our experts. AP refers to Autopilot controlled maneuvers, and NN refers to maneuvers controlled by our neural network expert. The SGDA method outperforms all baselines in learning a more semantically accurate imitation.}
     \label{fig:outcomematch_results}
\end{figure}

\begin{figure}
     \centering
\includegraphics[width=0.8\linewidth]{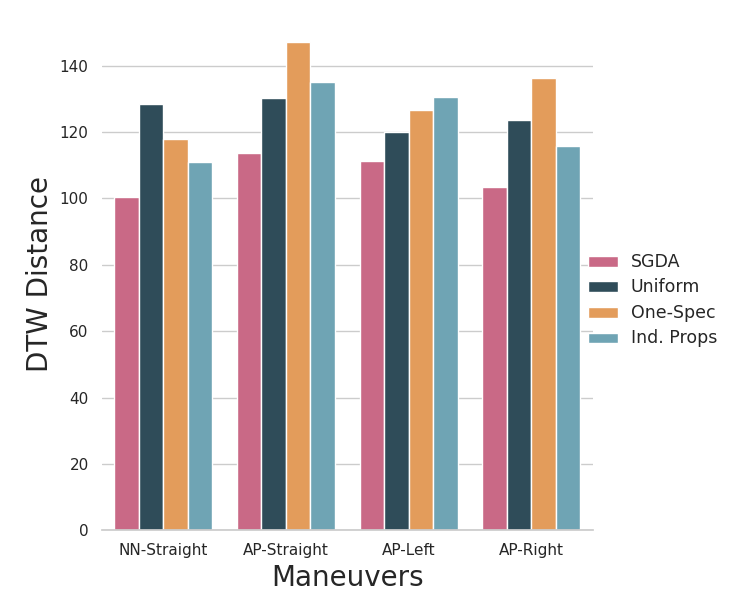}
     \caption{Results for Trajectory-Length Distances for imitations of the maneuvers performed by our experts, measured by Dynamic Time Warping (DTW) distance. The SGDA method outperforms all baselines in producing trajectories that are most similar to the expert.}
     \label{fig:dtw_results}
\end{figure}

In this section, we evaluate performance against our baselines to see if SGDA produces a more semantically similar imitation to our expert (Fig.~\ref{fig:outcomematch_results} and Table~\ref{tab:outcome_match_percentages}), if trajectories generated by SGDA resemble the experts' more closely (Fig.~\ref{fig:dtw_results}), and if SGDA achieves competitive test-set loss (Table~\ref{tab:loss_values}).

We evaluated the performance of each method on a test dataset collected from each expert in each maneuver. In collecting a test set, we aimed to not overrepresent unlikely outcomes in our dataset, while ensuring that every outcome was at least marginally represented. So as to not bias the test dataset towards SGDA, which explicitly samples for underrepresented specifications, we collected a dataset of 500 trajectories from the expert for each experiment, sampling from a uniform distribution over all environment parameters. To ensure a minimal representation of each specification, we first performed rejection sampling for each specification until trajectories that satisfied a given specification represented at least 
20 trajectories (four percent of the overall dataset).


\begin{table*}[]
    \normalsize
    \centering
    \caption{Number of samples for each specification in the expert test dataset and outcome match rates for each baseline in the NN-expert straight maneuver experiment. SGDA outperforms baselines in most outcome-specific accuracy and achieves comparable accuracy in all outcomes.}
    \label{tab:outcome_match_percentages}
    \begin{tabular}{ccc|c|cccc}
        \toprule
        (1) Avoid & (2) Do not & (3) Do not & & \multicolumn{4}{c}{\textbf{NN-Straight Match Rates}} \\
        Collisions & Halt & Abruptly Brake & \# Outcomes & SGDA & Ind. Props & Single-Spec & Uniform \\
        \midrule
        \cmark & \cmark & \cmark & 197 & 94\% & 96\% & \textbf{97\%} & 95\% \\
        \cmark & \cmark & \xmark & 20 & \textbf{50\%} & 40\% & 0\% & 0\% \\
        \cmark & \xmark & \cmark & 44 & \textbf{52\%} & 41\% & 14\% & 23\% \\
        \cmark & \xmark & \xmark & 65 & 78\% & 62\% & \textbf{86\%} & 57\% \\
        \xmark & \cmark & \cmark & 20 & \textbf{95\%} & 85\% & 70\% & 50\% \\
        \xmark & \cmark & \xmark & 20 & \textbf{45\%} & 0\% & 0\% & 0\% \\
        \xmark & \xmark & \cmark & 94 & \textbf{80\%} & 61\% & 79\% & 66\% \\
        \xmark & \xmark & \xmark & 40 & \textbf{23\%} & 5\% & 0\% & 0\% \\
        \bottomrule
    \end{tabular}
\end{table*}


\textbf{Evaluative metrics analysis.} In Fig.~\ref{fig:outcomematch_results}, we present results on outcome matching for imitation learned models using SGDA and our baseline methods of environment sampling. In each maneuver and with each model, models trained with SGDA have a higher rate of outcome matching than all of the baseline methods. Intuitively, we can interpret this result as SGDA-learned models being overall more \textit{semantically accurate} with respect to the expert than baselines. Moreover, in Fig.~\ref{fig:dtw_results}, we present the results comparing SGDA to our baselines in measuring the average trajectory-length distance between a learned model and expert pair of trajectories, collected in identical environment conditions. We observe that SGDA-learned models consistently have a lower average trajectory-length distance, supporting the notion that SGDA leads to models that not only semantically imitate the expert more closely, but have closer imitations in a (simulated) physical sense as well. As a final comparison, we compute a test set loss for each model, presented in~\ref{tab:loss_values}. Although this does not capture semantic similarities nor potential errors compounded during a trajectory-length execution, we note that SGDA-learned models have comparable (and marginally lower) test losses than baseline methods. 

\begin{table}[]
    \centering
    \caption{L1 Test Loss values for each baseline on our expert test dataset for each experiment. SGDA achieves consistently lower L1 errors.}
    \label{tab:loss_values}
    \begin{tabular}{c|cccc}
        \toprule
        \textbf{Test Loss} & NN-Straight & AP-Straight & AP-Left & AP-Right \\
        \midrule
        SGDA & \textbf{.059} & \textbf{.044} & \textbf{.066} & \textbf{.052} \\
        Uniform & .072 & .049 & .068 & .054 \\
        Single-Spec & .067 & .067 & .071 & .061 \\
        Ind-Props & .063 & .057 & .083 & .056 \\
        \bottomrule
    \end{tabular}
    \vspace{2mm}
\end{table}

Given the improvements in imitation displayed by SGDA-learned models, we seek to understand what semantic outcomes most strongly contribute to these improvements. To measure this, we compute the rate of outcome matching for each possible specification (outcome) in our test dataset, and compare SGDA against our baselines. 

\textbf{Outcome-specific analysis.} In Table~\ref{tab:outcome_match_percentages}, we show the outcome matching rates for each specification achieved by the expert in our test dataset for the human-learned neural network imitation experiment. These results show that the improvements the SGDA-learned model enjoys over the baselines are most significantly attributed to less likely outcomes. For example, outcomes where the expert collided and abruptly braked without halting ($\neg 1 \land 2 \land \neg 3$) were not matched at all by baseline methods, but our SGDA method was able to match nearly half of the outcomes. A similar story is the case with outcomes where the expert collided and came to a halt while abruptly braking ($\neg 1 \land \neg 2 \land \neg 3$). Being able to properly match these outcomes, which are highly correlated with other outcomes (i.e., you must brake in order to halt), requires being able to imitate nuanced behavior exhibited by the expert. 

In contrast, SGDA-learned models have a comparable, but often slightly lower, accuracy in more likely outcomes, such as outcomes where the expert did not collide, halt or abruptly brake $( 1 \land 2 \land 3)$, or the case where the expert does not collide but halts and abruptly brakes $( 1 \land \neg 2 \land \neg 3)$. This difference in performance can be directly attributed to imbalances in the training data for each baseline. For our single-spec baseline sampling, for instance, the method aimed to falsify just one specification, and found a large number of environment conditions that did so in very similar ways (specifically, finding conditions that satisfied $( 1 \land \neg 2 \land \neg 3)$). In fact, of the expert training data for the single-spec baseline, 38\% of the data satisfied $( 1 \land 2 \land 3)$, and 53\% satisfied $( 1 \land \neg 2 \land \neg 3)$. In contrast, for SGDA's training data, the percentages for each of those specifications were 35\% and 24\%, respectively. Overall, our results indicate that SGDA will greatly help improve imitated behavior in unlikely but semantically meaningful outcomes, without significantly altering the accuracy in behavior in common outcomes.

\textbf{Property robustness analysis.} Lastly, to provide insight into SGDA's robustness to changing properties in the set $LP$, we repeat our NN-Expert experiment with varying thresholds for the braking value in our ``Do not abruptly brake'' property. We notice that the rarity with which the property occurs indeed has an effect on the ability of SGDA to imitate expert behavior on that property. However, we note that SGDA outperforms our property-agnostic baseline -- uniform sampling -- even when the property becomes rare, and is generally robust to changes when the property is more frequent. We provide full details of this experiment in the appendix.

\section{Conclusion} 
\label{sec:conclusion}


In this paper, we present a method that leverages task-relevant information to systematically sample new environments for data aggregation in the context of imitation learning. Within this method, we introduce the notion of a \textit{Semantic Trajectory Partition} that divides up a space of possible environments into elements that each represent semantically different outcomes for an agent. Our experiments show that aggregating data using our method outperforms existing methods of environment sampling, especially in cases that are unlikely but semantically meaningful to a user. 

There are many possible avenues for future work. One could combine SGDA with other DAgger-style approaches to further improve learned imitations by controlling both \textit{when} and \textit{where} expert data should be collected. Another direction is to use SGDA to learn imitations in a human modeling setting, especially for individuals going through training in simulation, to better understand their areas of improvement. 

\section*{Acknowledgments}
Toyota Research Institute provided funds to support this work. A. Shah is supported by the Office of Naval Research under an NDSEG Fellowship.

\clearpage

\bibliographystyle{plainnat}
\bibliography{references}
\clearpage

\section*{Appendix}
\subsection*{Additional Experimental Details and Results}

\textbf{Details on experiment implementation}
As mentioned in the main text, we use Bayesian Optimization with Gaussian processes as our choice of optimization-guided sampler. When we sample using one of these optimizers, we have it maximize a target value that is equivalent to the quantitative semantics of the STL specification it is sampling for. In the case of the individual properties baseline, these specifications are simply $G(p)$, where $p \in LP$ are the individual properties themselves, and $G$ is the globally operator in temporal logic semantics. To encourage exploration, we add a term to the optimizer's target value that adds the following quantity \textit{if and only if} the sample and its corresponding trajectory satisfies the specification:
\[
a \cdot \min_{\xi_{\pi} \in \varphi}(\text{distance}(\xi_{\pi}, \xi_{\pi}^{c}))
\]
Here, $\xi_{\pi}^{c}$ is the current trajectory of the sampled environment condition that is being evaluated, and $a$ is our exploration hyperparameter. Intuitively, we find the \textit{minimum} distance between our current trajectory and all other trajectories that satisfied the same trajectory, and use that as our exploration bonus. Although this does not guarantee a minimum level of exploration, it does incentivize finding new regions of the environment condition space that also satisfy our specification. We keep the same exploration bonus with the same hyperparameter multiplier $a$ for all baselines that use optimization-based sampling.

For our behavior cloning models, we use a 3-layer, 256-unit MLP network as our model class. We train the model for 100 epochs on its training dataset with a learning rate of 0.0001 and a batch size of 500. We perform all training on an NVIDIA RTX A4500, which is the same hardware on which CARLA environments were sampled and simulated on. Our training data was a vector of features that captured relevant information about the trajectory and environment, namely, the ego's location, velocity, heading, and acceleration, and if the ado was within the line of sight (i.e., not blocked by the occlusion in the environment), the ado's location and speed, both modified by injecting Gaussian noise (centered at zero in both cases, with variances of two and one for location coordinates and speed, respectively.) 

\textbf{Additional Results.} In Tables~\ref{tab:ocm_ap_straight},~\ref{tab:ocm_ap_left}, and~\ref{tab:ocm_ap_right}, we present the results for outcome-specific matching rates for each of the CARLA autopilot maneuvers. Like the results for our neural network expert, we see that SGDA largely outperforms our baselines in having higher outcome-specific matching rates, especially in cases where the outcomes are highly unlikely. On more likely outcomes, SGDA is able to perform close to the level of our baselines, which have seen significantly more of these outcomes for, such as the example mentioned in the main text. Part of these discrepancies can be attributed to the limited amount of data shown to each model during training. We expect that with more data, SGDA will improve even more relative to baselines, since SGDA will aggregate data on outcomes that it matches poorly on (both unlikely and likely), whereas the baseline methods do not have a principled way to sample many different unlikely outcomes and improve their imitations in such settings.

\textbf{Robustness Experiment.} As mentioned in the main text, we run a simple experiment to observe how the performance of SGDA changes when we vary a property in the set $LP$. Specifically, we repeat our NN-Expert experiment, and vary the threshold for the braking threshold in our ``Do not abruptly brake'' property, keeping the other properties fixed. By varying this threshold, we change the frequency with which this property is violated by our expert in different environment conditions. We run SGDA for two rounds, with the same additional properties (avoid collisions and do not halt) and experimental setup described in the main text. We then measure the matching percentage of SGDA on \emph{just} the property we vary. Concretely, we observe: if the expert violated the braking property when set to a certain threshold, does our learned imitation violate the property at that threshold as well? We study the performance on this property in isolation from the others.

\begin{figure}
     \centering
\includegraphics[width=0.98\linewidth]{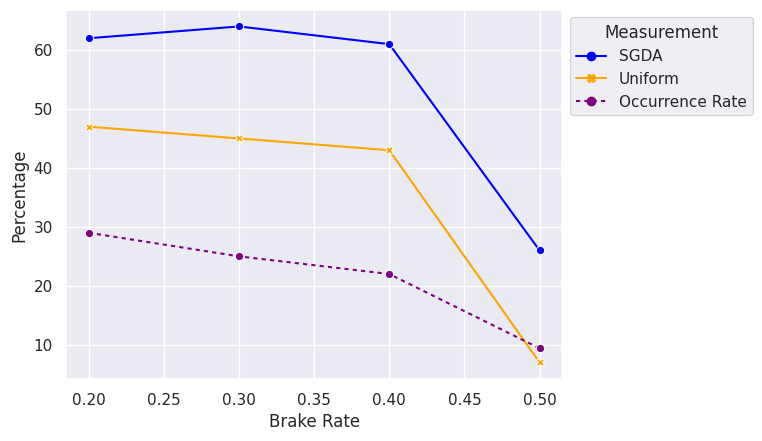}
     \caption{Matching rates (in orange and blue) for SGDA and the uniform baseline on violations of the brake property, when varying the brake rate threshold. We also present the frequency (in purple) with which the property is violated when the threshold is set at varying rates.}
     \label{fig:robustness}
\end{figure}

We present our results in Fig.~\ref{fig:robustness}. We collect a test dataset of 500 episodes, sampling from the uniform prior distribution of environment conditions (without any rejection sampling, as done in the main experiment.) The purple (dashed) line records how frequently the property is violated in the expert trajectories in our test dataset, at a given braking threshold. We see that as the braking threshold (x-axis) increases, the frequency of observing a violation decreases slightly, then drops off significantly from a threshold of 0.4 to 0.5. The blue line shows SGDA's  matching percentage for \textbf{violations} of this single property. To compare with a baseline, we present the same results for the uniform baseline, shown in orange. We see that SGDA's matching rate changes little for threshold values from 0.2-0.4, even as the frequency drops from just below 30 percent to just above 20 percent. However, at a threshold of 0.5, when the frequency drops to below 10 percent, both SGDA and the uniform baseline's performances drop precipitously. This indicates to us that the frequency of seeing a property has a strong impact on SGDA's performance, when that frequency falls below a certain threshold. 

We reason that both a larger pool of environments to sample and select from, as well as more rounds of data aggregation, will help mitigate such dropoffs in practice. Measuring the relative frequencies of properties included in $LP$ will help inform the choices of hyperparameters when using SGDA, such as the amount of data to aggregate, how many environments to sample and select, and how to balance exploitation and exploration in the UCB algorithm during environment sampling. The exact nature of the property matters as well - we observe that in situations where our NN-expert abruptly brakes with a high brake rate, that high rate is only applied for a small number of initial timesteps before the rate comes down and the expert slows down more gradually. Our imitations had trouble replicating this exact behavior during learning, which further explains the dropoff in performance as the threshold grew to 0.5.

\begin{table*}[]
    \centering
    \normalsize
    \caption{Number of samples for each specification in the expert test dataset and outcome match rates for each baseline in the Autopilot straight maneuver experiment.}
    \label{tab:ocm_ap_straight}
    \begin{tabular}{ccc|c|cccc}
        \toprule
        (1) Avoid & (2) Do not & (3) Do not & & \multicolumn{4}{c}{\textbf{AP-Straight Match Rates}} \\
        Collisions & Halt & Abruptly Brake & \# Outcomes & SGDA & Ind. Props & Single-Spec & Uniform \\
        \midrule
        \cmark & \cmark & \cmark & 226 & 91\% & 85\% & \textbf{92\%} & \textbf{92\%} \\
        \cmark & \cmark & \xmark & 20 & \textbf{15\%} & 0\% & 0\% & 0\% \\
        \cmark & \xmark & \cmark & 67 & \textbf{92\%} & 76\% & 21\% & 34\% \\
        \cmark & \xmark & \xmark & 65 & 88\% & \textbf{91\%} & 75\% & 44\% \\
        \xmark & \cmark & \cmark & 20 & \textbf{80\%} & \textbf{80\%} & 55\% & 50\% \\
        \xmark & \cmark & \xmark & 26 & \textbf{31\%} & 23\% & 15\% & 0\% \\
        \xmark & \xmark & \cmark & 20 & \textbf{70\%} & 50\% & 20\% & 0\% \\
        \xmark & \xmark & \xmark & 56 & \textbf{20\%} & 0\% & 0\% & 7\% \\
        \bottomrule
    \end{tabular}
\end{table*}

\begin{table*}[]
    \normalsize
    \centering
    \caption{Number of samples for each specification in the expert test dataset and outcome match rates for each baseline in the Autopilot left-turn maneuver experiment.}
    \label{tab:ocm_ap_left}    
    \begin{tabular}{ccc|c|cccc}
        \toprule
        (1) Avoid & (2) Do not & (3) Do not & & \multicolumn{4}{c}{\textbf{AP-Left Match Rates}} \\
        Collisions & Halt & Abruptly Brake & \# Outcomes & SGDA & Ind. Props & Single-Spec & Uniform \\
        \midrule
        \cmark & \cmark & \cmark & 191 & 82\% & \textbf{89\%} & \textbf{89\%} & 88\% \\
        \cmark & \cmark & \xmark & 20 & \textbf{20\%} & 0\% & 0\% & 0\% \\
        \cmark & \xmark & \cmark & 115 & 92\% & \textbf{96\%} & 95\% & 87\% \\
        \cmark & \xmark & \xmark & 20 & \textbf{65\%} & 0\% & \textbf{65\%} & 10\% \\
        \xmark & \cmark & \cmark & 20 & \textbf{35\%} & \textbf{35\%} & 10\% & 0\% \\
        \xmark & \cmark & \xmark & 68 & \textbf{37\%} & 0\% & 13\% & \textbf{37\%} \\
        \xmark & \xmark & \cmark & 20 & \textbf{90\%} & 40\% & 5\% & 0\% \\
        \xmark & \xmark & \xmark & 46 & \textbf{70\%} & 0\% & 0\% & 20\% \\
        \bottomrule
    \end{tabular}
\end{table*}

\begin{table*}[]
    \normalsize
    \centering
    \caption{Number of samples for each specification in the expert test dataset and outcome match rates for each baseline in the Autopilot right-turn maneuver experiment.}
    \label{tab:ocm_ap_right}
    \begin{tabular}{ccc|c|cccc}
        \toprule
        (1) Avoid & (2) Do not & (3) Do not & & \multicolumn{4}{c}{\textbf{AP-Right Match Rates}} \\
        Collisions & Halt & Abruptly Brake & \# Outcomes & SGDA & Ind. Props & Single-Spec & Uniform \\
        \midrule
        \cmark & \cmark & \cmark & 195 & 88\% & \textbf{93\%} & 91\% & \textbf{93\%} \\
        \cmark & \cmark & \xmark & 20 & \textbf{20\%} & 5\% & 0\% & 0\% \\
        \cmark & \xmark & \cmark & 97 & \textbf{73\%} & 67\% & 34\% & 60\% \\
        \cmark & \xmark & \xmark & 20 & \textbf{30\%} & 0\% & 0\% & 0\% \\
        \xmark & \cmark & \cmark & 20 & \textbf{80\%} & 50\% & 55\% & 70\% \\
        \xmark & \cmark & \xmark & 55 & \textbf{62\%} & 25\% & 18\% & 16\% \\
        \xmark & \xmark & \cmark & 37 & \textbf{84\%} & 41\% & 79\% & 38\% \\
        \xmark & \xmark & \xmark & 56 & 28\% & 34\% & \textbf{39\%} & 23\% \\
        \bottomrule
    \end{tabular}
\end{table*}

\subsection*{Additional Algorithmic Details} 

\textbf{Property Weights.} In section~\ref{sec:env_partition}, we noted that a potential issue of our STP method is the partition size that is exponential in the size of our property set, $p \in LP$. If we were to run the original algorithm on an extremely large partition, we would likely be sampling for specifications that we may value less than others, leading to inefficiency in environment sampling and selection. To combat this, we generalize SGDA and our STP formation to a setting where a user can provide \textit{weights} for each property in their set to quantify how semantically meaningful the existence of a given property is relative to other properties. 

Formally, we define a \textit{weighted} property set as a tuple $(w, p) \in LP^w$, where each weight $w$ associated with a property is a real-valued number in the range $(0, 1)$. We can then take the product of these weights (or their complements, in the case of negated properties), to assign weights to entire specifications. To make this more concrete, recall the simple example introduced in section~\ref{sec:env_partition}, where we have just two properties: \textbf{A} (avoid collisions) and \textbf{B} (avoid speeding.) Suppose that a user is very interested in outcomes where collisions occur, and only slightly interested in seeing outcomes where the car is not speeding. Following this reasoning, the user assigns the following property weights: \textbf{A} is assigned a weight of 0.1, and \textbf{B} is assigned a weight of 0.6. This means that the negation of \textbf{A} (in presence of a collision) is assigned the complementary weight to \textbf{A}, or (1 - 0.1) = 0.9, and the negation of \textbf{B} is assigned (1 - 0.6) = 0.4. These weights then are multiplied to form weights for each specification in the STP. In this example, $(A \land B)$ would have weight $(0.1 * 0.6) = 0.06$, $(\neg A \land B)$ would have $(0.9 * 0.6) = 0.54$, $(A \land \neg B)$ would have $(0.1 * 0.4) = 0.04$, and $(\neg A \land \neg B)$ would have $(0.9 * 0.4) = 0.36$. 

Once we have computed weights for each specification in our STP, leveraging these weights in the SGDA algorithm is conceptually straightforward. We can use each specification's weight as a multiplicative factor in the Q-value assigned to each specification defined in algorithm~\ref{ecs_alg}. These weighted Q-values will then be used by the UCB algorithm to select future specifications to sample for. This means that specifications with lower weights will have lower Q-values, and are less likely to be sampled for during the sampling step of SGDA.

\textbf{Computational Overhead of SGDA.} 
The SGDA algorithm will incur an asymptotic computational overhead with respect to the hyperparameters of the algorithm that differs from other environment sampling approaches. We informally discuss the overhead here.

The computational overhead that SGDA sustains lies primarily in the sampling step (EC-Samp). In the sampling step, outlined in algorithm~\ref{ecs_alg}, the following computations will add nontrivial overhead: (1) the invoking of an optimization-based sampler to get new environment conditions, (2) actuating the dynamics to collect a trajectory $ \xi_{\pi^{\text{IL}}}(e) = d(\pi^{\text{IL}}, e) $ and checking which specification in the partition $(e, \xi_{\pi^{\text{IL}}}(e))$ falls into, and (3) computing the Q-values for each specification and invoking the UCB computations with these Q-values to select the next specification to sample for. 

Regarding (1), distributions learned by optimization-guided samplers are generally learned iteratively and online, so this computation will contribute only minor overhead. Regarding (3), the computational complexity for these operations will scale linearly with the size of the partition, but the computation of Q-values and UCB values for each specification only requires simple arithmetic operations and will not heavily burden the computation. Both algorithmically and in practice, we find that (2) adds the most overhead: invoking the dynamics for a given environment condition can be expensive and is done for each sample. Monitoring each specification to determine the $\varphi \in \Phi$ for a given $(e, \xi_{\pi^{\text{IL}}}(e))$ pair satisfies also requires evaluating the semantics of every $\varphi$ in the worst-case. 

We note that the combination of these operations contributes nontrivial, but not prohibitive, overhead in practice, and that this overhead will scale linearly with the size of $\Phi$ and exponentially with the size of $LP$. However, many parts of this overhead are also shared by other environment sampling techniques (such as our two baselines), namely in invoking an optimization-guided sampler and actuating the dynamics. More importantly, as evidenced by our experiments, we trade off the increase in computation for an increase in model performance, and therefore less required involvement of an expert model. Such improvements are critical when expert models are expensive to query, such as in human studies, which we see as a primary contribution of SGDA.

\end{document}